**Assessing instructor-AI cooperation for grading essay-type questions in an introductory sociology course**

Francisco Olivos[a,b]

Tobias Kamelski[a]

Sebastián Ascui-Gac[a]

[a]Department of Sociology and Social Policy, Lingnan University

Conflict of interest: None

[b]Corresponding author. Email: franciscoolivosrave@LN.edu.hk

**Assessing instructor-AI cooperation for grading essay-type questions in an introductory sociology course**

**Abstract**

This study explores the use of artificial intelligence (AI) as a complementary tool for grading essay-type questions in higher education, focusing on its consistency with human grading and potential to reduce biases. Using 70 handwritten exams from an introductory sociology course, we evaluated generative pre-trained transformers (GPT) models' performance in transcribing and scoring students' responses. GPT models were tested under various settings for both transcription and grading tasks. Results show high similarity between human and GPT transcriptions, with GPT-4o-mini outperforming GPT-4o in accuracy. For grading, GPT demonstrated strong correlations with the human grader scores, especially when template answers were provided. However, discrepancies remained, highlighting GPT's role as a "second grader" to flag inconsistencies for assessment reviewing rather than fully replace human evaluation. This study contributes to the growing literature on AI in education, demonstrating its potential to enhance fairness and efficiency in grading essay-type questions.

**Keywords:** AI-assisted grading, higher education, bias reduction, essay-type assessment, generative pre-trained transformers.

**Introduction**

By solving kinds of problems previously reserved for humans (McCarthy et al. 1955), educational experts have asserted that artificial intelligence (AI) has revolutionized education (Popenici and Kerr 2017; Wang et al. 2024). Recent figures indicate that 86% of higher education students from 16 countries regularly utilize AI in their studies, with 58% declaring that they feel unconfident with their knowledge and skills (Digital Education Council 2024). If not a revolution, the popularity and increasing pervasiveness of AI over recent years have been highly consequential for higher education.

In this context, AI has impacted academia in its dimensions of teaching, research, and service (Barros, Prasad, and Śliwa 2023). A primary concern raised by students' use of AI has been cheating (Ratten and Jones 2023), prompting challenges on the traditional methods of assessment. Moreover, several guides have provided a landscape of alternatives for effectively integrating AI into teaching (Levy and Pérez Albertos 2024). For instance, Olivos and Liu (2024) have demonstrated how students in research methods courses can use generative pre-trained transformers (GPT) models to pilot their questionnaires as an initial stage in survey design. Lee and Yeo (2022) developed an AI-based chatbot to simulate interactions with a student struggling with math misconceptions, helping teachers practice responsive teaching skills. In 2023, Google Labs released NotebookLM, enabling users to create podcasts based on documents, which could serve as a means of disseminating students' creative work. Thus, AI promises to automate, enhance, or facilitate entire or partial aspects of teaching.

One particular application of AI in teaching is in the process of grading class assignments or exams, or what Wang and colleagues (2024) term "intelligent assessment." Previous estimates suggest that 40% of teaching time is spent on grading and providing feedback (Mandernach and Holbeck 2016). Consequently, if instructors are assisted by AI in these tasks, faculty members could be relieved of repetitive duties and focus on more advanced and complex activities.

Indeed, if we ask ChatGPT using GPT-4o model, a commonly used generative artificial intelligence, the following question: "*Can you grade essay-type questions if I provide the answers and a grading criteria?*," it confidently responds: "*Yes, absolutely! If you provide the answers and grading criteria, I can grade the essay-type questions for you*" (OpenAI 2024a).

However, recent studies have taken this a step further by actually examining the consistency between human and AI grading to critically evaluate its potential (e.g., Grivokostopoulou, Perikos, and Hatzilygeroudis 2017; Jukiewicz 2024; Wetzler et al. 2024).

Among the most promising evaluations, Grivokostopoulou et al. (2017) demonstrate a high level of agreement between an automated assessment system and human scoring in evaluating artificial intelligence curricula. Jukiewicz (2024) also reports positive results from ChatGPT models' evaluation of programming tasks but highlights the need for careful use and human teacher verification due to costs, potential hallucinations, and lack of reproducibility. In a closer application to sociology, Wetzler et al. (2024) compare human and AI grading of students' essays in an introductory psychology course. Their findings show no agreement between instructors and AI, with the latter delivering more lenient grades. Therefore, the promise of automatic scoring for students' assessments should be approached with caution.

Nevertheless, the literature indicates that human graders are also prone to biases. A large body of evidence documents teachers' racial or gender biases in student evaluation (Chin et al. 2020; Protivínský and Münich 2018), as well as the halo effect, in which previous experiences with a student influence subsequent evaluations (Malouff et al. 2014). Certainly, deidentification of assessments can help eliminate some sources of bias. However, other sources of biases can certainly be more persistent. Birkelund (2014) demonstrates inconsistent evaluations before and after lunch, similar to results found in judges' decisions (Danziger, Levav, and Avnaim-Pesso 2011). In another study, Klein (2003) shows that uninterrupted examination of a large number of tests over an extended period induces grading inconsistency. Therefore, rather than advocating for the replacement of human grading or assuming human grading as an infallible method, this study examines the consistency between human and AI grading to explore potential collaboration for a fairer grading process. Detecting inconsistencies in grading can mitigate potential biases introduced by instructors, and particularly inconsistent student work can be re-assessed. For this purpose, we compare human and GPT grading of 70 handwritten exams from an introductory sociology course, each containing 6 essay-type questions. GPT is a type of large language model (LLM) and a prominent framework in generative AI. It represents a subset of artificial intelligence specifically designed to process and generate human-like text based on patterns learned from vast amounts of data. As part of the broader AI ecosystem, GPT exemplifies how advanced machine learning models can move beyond mechanical tasks to perform complex, nuanced functions such as text generation, analysis, and evaluation.

Moreover, we advance the literature and pedagogical resources on the application of LLMs in education in several directions. First, we assess the human-GPT agreement in scoring essay-type questions, a format commonly used in humanities and social sciences. Previous studies have focused on coding tasks or essays (e.g., González-Calatayud, Prendes-Espinosa, and Roig-Vila 2021; Jukiewicz 2024; Wetzler et al. 2024). Essay-type questions are open-ended and evaluate students' understanding, critical thinking skills, and ability to articulate knowledge. However, as they rely on subjective grading, fairness becomes a more sensitive issue, and the cognitive processes required of AI go beyond purely mechanical and repetitive functions (Ratten and Jones 2023). Second, we use handwritten answers instead of computer-assisted assessments. Before comparing human and GPT assessments, we examine whether GPT transcriptions are consistent with human transcriptions for subsequent scoring. Early developments in other fields have demonstrated the practical potential of AI for scoring handwritten reading comprehension essays (Srihari et al. 2008). Third, we make available the Python script used to transcribe the handwritten answers and to implement the scoring system (see data availability statement).

In the following section, we describe the data and methodology of the study. In the second section we report the results of the analysis and, in the final section, we provide conclusion and a discussion of the challenges and opportunities of instructor-AI cooperation for scoring essay-type questions.

## 1. Data & Methods

*Participants*

*Human coder*. A research assistant photographed each of the 6 answers in separate files for each of the 70 students of the course Introduction to Sociology at [ANONYMIZED]. The research assistant then transcribed the responses into an Excel spreadsheet.

*Human instructors*. An instructor (Grader A) and two teaching assistants (Graders B and C) were responsible for scoring the exams based on a standard template for answers. The primary instructor, who also delivered the lecture content of the course, graded the first four essay questions. Each teaching assistant then graded one of the remaining two questions. The questions are included in the supplementary material.

*AI Grading Systems*. For transcribing the photographed answers, we utilized the application programming interface (API) equivalent of ChatGPT (OpenAI API) in Python. GPT-4o and GPT-4o-mini were utilized for comparison purposes. As previous models are not able to process visual data, the grading was performed using the models GPT-4o and GPT-4o-mini as discussed in the analysis section.

The study was reviewed and approved by the Sub-Committee on Research Ethics and Safety under the Research Committee at [ANON] (approval reference: ECO26-2425), ensuring adherence to ethical standards in research.

*Procedures*

Each answer was photographed by the human coder using a CMOS Leica Vario-Summilux 1-16/2.2 r4 75 asph camera. The camera was set to the "2x Leica vibrant" mode and positioned at a fixed distance for each photo. We utilize this device because this is a standard camera integrated to a cellphone, which may be available for most of the instructors that would like to adopt this grading system.

The human coder was instructed to write down each of the answers into an Excel spreadsheet, ordered as a data matrix, transcribing the verbatim to the best of their ability. They did not correct any grammar mistakes, typos, or incomplete sentences. Unreadable words were skipped. A total of 420 photos were taken, equivalent to 70 exams of 6 questions each.

As the first step of the analysis, human transcriptions were compared to the transcriptions generated by GPT-4o and GPT-4o-mini. GPT-4o is designed for high versatility outputs, while GPT-4o-mini is a smaller, faster, and more cost-efficient alternative. While GPT-4o is generally expected to deliver better performance due to its advanced capabilities, GPT-4o-mini could potentially outperform it in this task, as its simpler and more straightforward processing might align more closely with the goal of literal transcription. This makes GPT-4o-mini a cost-effective option, particularly for widespread adoption in resource-constrained environments. To ensure consistency, the temperature was set to 0.3 for both models, promoting direct transcription of the text without creative elaboration. Additionally, two different prompts were utilized to guide the transcription process for each model.

The GPT models cannot directly process common image formats like .jpg. To make these images compatible, they must be converted into a format the GPT models can understand. This process is automated in the ChatGPT service but requires manual conversion when using the OpenAI API. We encoded each photographed answer using the Base64 Python package, an encoding method that converts binary data, such as digital images, into ASCII strings that GPT models can process. Once encoded, the ASCII strings are passed to GPT models using the prompts provided below.

First, we requested the GPT models to transcribe the answers exactly as written (see Figure 1).

[Figure 1 about here]

The second prompt allows the GPT models to use the best of their ability, allowing to utilize their best guess if words are unclear or unreadable. The prompt is displayed in Figure 2.

[Figure 2 about here]

Next, to answer the question whether the GPT models can accurately transcribe the students' handwritten answers, we estimated the similarity between the human transcription and both transcriptions by done by the GPT models. Cosine similarity index is a commonly used measure in computational text analysis to determine how similar are two pieces of text because it is less sensitive to word count, space usage or exact word positioning (for an accessible explanation, see Stoltz and Taylor 2024). This technique treats each document as a vector in a multidimensional space, where each dimension corresponds to a unique word from the combined vocabulary of both documents. In this case, a particular answer transcribed by the human coder and the corresponding answer transcribed by the GPT models, as the values in these vectors represent the frequency of each word in the documents.

To understand cosine similarity, imagine each document as a list of how often each word appears. These lists are converted into vectors. Cosine similarity measures the cosine of the angle between these two vectors. An angle close to 0 degrees shows that the documents are very similar, meaning the cosine similarity is close to 1. Conversely, an angle close to 90 degrees indicates that the documents are dissimilar, making the cosine similarity close to 0. In the context of this study, a cosine similarity scores close to 1 would suggest that the GPT models produced a transcription that is comparable to the human transcription. Conversely, a score near 0 would indicate a considerable discrepancy between the two transcriptions. The stringdist package in R was utilized to estimate Cosine similarity (van der Loo 2014), calibrating the index computation by choosing a N-gram size of 3, which improves the accuracy of the comparison when dealing with long strings of text.

In the second stage, and as suggested by the similarity analysis below, we compared the scores of human graders and GPT-4o and GPT-4o-mini. Two prompts were utilized to evaluate the consistency with the human scores. As shown in Figure 3, in the first scoring prompt, we instructed the GPT models to use their own knowledge with a series of specific instructions of the grading system.

[Figure 3 about here]

A second prompt follows the same structure but provides template answers to each of the questions as a parameter for evaluation. The prompt is displayed in Figure 4.

[Figure 4 about here]

We also manipulated the temperature. The temperature in the OpenAI API refers to a parameter used to control the randomness of the responses generated by the model, ranging from 0 to 2. Lower values produce more deterministic or predictable responses. Higher values increase randomness, diversity, and creativity in the responses. The default value for the temperature is 1.0. However, 1.0 is already considered to produce creative outputs, with OpenAI API deeming 0.8 a high value (OpenAI 2025). We selected 0.7 and 0.2 as the upper and lower ends, avoiding the practical and technical extremes. Thus, we have a total of four settings for each prompt: (1) GPT-4o-mini at temperature 0.7, (2) GPT-4o at temperature 0.7, (3) GPT-4o-mini at temperature 0.2, and (4) GPT-4o at temperature 0.2.

For each setting, we instructed the models to grade the answers 100 times to evaluate the consistency of the results. Then, the outputs were averaged to utilize one single score for each student using each setting. Two types of analyses are conducted to compare the human benchmark and the GPT scores. First, we estimated the Pearson's correlation between the human scores and the averaged score produced by each setting and prompt. Then, we utilized Bland-Altman analysis (1999), which is a method of data plotting used in analytical chemistry or biomedicine to analyze the agreement between two different assays. Bland-Altman plots are extensively used to evaluate the agreement between two instruments or measurement techniques. Recently, they have been applied to compare human and AI scores (Wetzler et al. 2024). This method involves plotting the differences between two measuring techniques against the averages of the same measurements, allowing identification of any systematic difference (fixed bias) or possible outliers. This analysis acknowledges that even the "gold standard" of human grading is not without error but serves as a useful benchmark for comparison.

**Results**

*Image transcription*

Table 1 displays the mean Cosine similarity index and its standard deviation between human and GPT models' transcriptions, which utilize a prompt for literal transcription (1a) or a prompt enabling the best guess (2a). The mean and standard deviation of the Cosine similarity for each pair of texts are reported for each question together with their overall mean. The mean indicates that GPT-4o-mini provides transcriptions slightly more similar and with a smaller standard deviation than GPT-4o. Nevertheless, in both cases, the transcriptions exhibit a high level of similarity to the human coder's transcriptions.

[Table 1 about here]

When comparing the different prompts, the average indicates that enabling GPT-4o to use its best guess does not increase the similarity with the human coder. Similarly, the average for GPT-4o-mini does not exhibit differences between prompts. However, when looking at the particular questions, the prompt enabling the best guess shows slightly higher similarity from question 1 to question 5 but not question 6. Even though we cannot ensure whether the human transcriptions totally mirror the answers written by the students, the overall analysis suggests that GPT can successfully mirror the human transcriptions, and GPT-4o-mini slightly outperforms GPT-4o .

*Scoring prompt (1b) utilizing GPT models' sociological knowledge*

We requested GPT-4o-mini and GPT-4o to evaluate the transcribed students' responses. Noteworthy, GPT-4o-mini is considerably more cost-effective (OpenAI 2024b), which offers a lower cost alternative for educators. We utilize the transcriptions by GPT-4o-mini. As described above, the first prompt (1b) requested the scoring of the responses without template responses. It allows the GPT models to use their own knowledge to score the assessments.

For each GPT model and at different temperatures, we asked the GPT models 100 times to score. The procedure enables us to examine the consistency of results. Lower temperatures generate more deterministic results; therefore, we may expect lower variability in the outputs produced by settings at temperature 0.2. Figure 5 displays the standard deviation across 100 runs of the same task under different settings of model and temperature for the GPT models. The data show variability in the consistency of results, which is not uniform across different models or temperature settings. This is explained because of the stochastic nature of the process. Specifically, GPT-4o-mini provides more consistent results, particularly at a lower temperature setting (0.2), which is indicative of its more aggressive generalization and conservative scoring approach. The higher kurtosis also indicates that scores tend to concentrate around the mean in GPT-4o-mini set up at temperature 0.2. This behavior is possibly due to differences in model size and training data between the models. Lower temperatures typically reduce randomness in the model's outputs, resulting in closer adherence to the most likely response pattern, which may explain the reduced variability seen in GPT-4o-mini at temperature 0.2. GPT-4o-mini is trained to generalize more aggressively (OpenAI 2024b), which may lead to more consistent and conservative scoring, i.e., smaller range of scores and, hence, a smaller standard deviation.

[Figure 5 about here]

To assess whether the human and GPT scoring are consistent across settings, we estimated the correlation between the human scoring and the average of the 100 tasks for each combination of model and temperature. These correlations are displayed in Figure 6.

[Figure 6 about here]

Figure 6 suggests a positive correlation between the human scoring for the different settings. It indicates that in general, GPT can meaningfully utilize its own knowledge to assess students' answers. However, the correlation can be deemed moderate at conventional levels of correlational strength. When comparing the models, GPT-4o-mini set at the default temperature 0.7 and 0.2 exhibits the worse performance when using the human scoring as benchmark. Conversely, the correlations between the benchmark human scoring and scores from the more complex models are substantially higher, particularly when it is set at temperature 0.7.

Bland-Altman graphs are displayed in Figure 7. These figures are instrumental to visualize the agreement between the human and automated scores across different settings. On the x-axis, each plot represents the average of human and GPT scores for each student, while the y-axis measures the score difference, calculated as GPT's score minus the human grader's score. Points closer to the zero line on the dotted axis indicate a higher agreement, showing minimal variance between the two scoring methods.

[Figure 7 about here]

Observations above the zero line where GPT consistently provided higher scores than human graders are particularly notable. Across the four plots, GPT generally tends to score more generously than the human graders, as evidenced by the positive average differences (blue dashed lines). This trend is more pronounced with the GPT-4o-mini model. While differences related to the temperature setting of the models are subtle, the standard temperature setting (0.7) for both GPT models shows slightly smaller differences, suggesting a closer alignment with human scoring. Overall, when configured at temperature 0.7, GPT-4o demonstrates the closest agreement with human scoring compared to other settings. However, the low level of correlation and the substantial difference in scoring does not provide robust evidence of potentially utilizing GPT for grading when template answers are not provided.

The green dashed lines represent the lower and upper bound of where 95% of the differences between the two measurements are expected to fall. These lines help to assess the extent of disagreement above the mean difference. Observations above the upper limit or below the lower limit are also considered outliers or significantly different. These outliers are particularly relevant for this exercise. It can be considered a formal standard for reassessment of any particular set of exams.

*Scoring prompt (2b) providing template answers*

The second prompt provided a template answer for each question to GPT. Thus, GPT does not need to rely completely on its knowledge, and clearer parameters of evaluation are provided. In other types of assessments, rubrics could also be included to define criteria and weights for each assessment. In the same vein as prompt 1b analyses, we requested 100 runs if the tasks for each setting combining models and temperatures.

Figure 8 displays the standard deviations for each run, grouped by settings of model and temperature. The histograms generally indicate variability in scoring consistency across all model and temperature settings. The kurtosis values provide insights to understand the distribution shapes. GPT-4o-mini at a lower temperature (0.2) exhibits higher kurtosis (8.64), indicating a more peaked distribution with values clustering around the mean, suggesting greater consistency in scoring compared to other settings. This is consistent with expectations that lower temperatures result in outputs with reduced variability, reflecting the model's restricted creative range at these settings. Conversely, GPT-4o at temperature 0.7 shows very low kurtosis (0.01), corresponding to a flatter distribution, which indicates a broader spread of scores and, thus, higher inconsistency. This pattern suggests that the smaller and simpler model combined with a lower temperature setting allows for lower freedom in response generation, leading to decreased variability in the automated scoring.

[Figure 8 about here]

We also examine the correlation with the total human scoring for each setting. Figure 9 reports these results. All the Pearson's correlations on or above .80 indicate a strong consistency between human scoring and GPT scoring. Among the different model and temperature set-ups, GPT-4o exhibits a higher level of consistency with human scoring. There is no variation between temperature 0.2 and temperature 0.7.

[Figure 9 about here]

Figure 9 also enables us to highlight the central contribution of GPT grading. It is clear there is a high level of consistency between human and GPT scoring. Nevertheless, the correlation is not perfect, and therefore, a replacement can lead to unfair evaluations. Thus, the results should not be interpreted as a suggestion for the wholesale implementation of GPT scoring. However, human scoring is also prone to bias as we explained above, and the residuals are the most useful information to produce more bias-free assessments. The residuals are the difference between the observations (dots in the figures) and the predicted value on the line. For example, in the first subfigure of GPT-4o mini with temperature set at 0.7, there is an outlier in the lower left corner. This student was assessed with a higher score by GPT and a lower score by the instructors. Therefore, this deviation is an indicator that the assessment may need to be revised again. This suggests a potential complementarity between human and AI rather than a replacement.

Figure 10 presents Bland-Altman plots comparing GPT scores to human scores across different models. The plots highlight the average differences between the human assessments and the GPT models at temperature values of 0.2 and 0.7. Although smaller than differences in prompt 1b, the data reveal distinct patterns of deviation in the scores, with notable differences between the human assessments and GPT-4o-mini. As shown by a concentration of observations in the upper side of the graphs, the human scoring tends to be lower than the scores of the GPT-4o-mini (blue dashed line). On the contrary, GPT-4o has higher agreement with human scoring as suggested by differences closer to zero, particularly at temperature 0.7 where the differences are slightly lower than temperature 0.2. Overall, these results indicate that while the GPT-4o-mini model tends to overestimate scores compared to human graders, GPT-4o maintains a closer alignment with human assessments, evidenced by smaller average differences. These results are also supported by correlational analysis in Figure 9. Similarly to the Bland-Altman plot utilizing prompt 1b, we can also identify exams for reassessment considering those exams under the lower limits.

[Figure 10 about here]

Given the noted superiority of prompt 2b over prompt 1b, we also examine the performance of individual graders based on the correlation and average difference with the GPT scores. Due to varying sources of bias, it is conceivable that different graders might evaluate essay-type questions distinctly. While a direct comparison of scores between human graders is impractical since they evaluated different questions, analyzing each question separately provides granular insights into GPT performance. Table 2 displays these results. Questions 1 through 4 were assessed by grader A, while questions 5 and 6 were evaluated by graders B and C, respectively. Notably, there exists significant heterogeneity in the correlation and average differences between human grades and GPT scores. The correlations range from 0.87 for question 2 with GPT-4o-mini (temperature 0.7) to 0.45 for question 6 with GPT-4o at both temperatures. There is also considerable variability within the questions assessed by grader A across different setups. This analysis underscores that while GPT can effectively aid human graders, the variability in outputs between graders and across specific questions is substantial. Consequently, GPT could serve as a supplemental tool or "second grader," but it in any case should not, or cannot, totally replace the human grading process.

[Table 2 about here]

**Conclusion**

Human grading is not free from biased outcomes. Previous studies have demonstrated that human graders exhibit various forms of bias when assessing students' assignments (Birkelund 2014; Klein and El 2003; Malouff et al. 2014; Malouff, Emmerton, and Schutte 2013; Protivínský and Münich 2018). In this study, we examined the potential of instructor-AI collaboration in the assessment of essay-type questions. Using handwritten exams from an introductory sociology course, we analyzed (1) the similarity between human and GPT image-to-text transcriptions and (2) the consistency between human and GPT scoring of students' answers. The results reveal a high degree of similarity between human and GPT transcriptions, with GPT-4o-mini slightly outperforming GPT-4o. Furthermore, the analysis indicates a high level of consistency between human graders and the GPT models when template answers are provided.

This study contributes to the growing literature on AI applications in teaching (Popenici and Kerr 2017; Wang et al. 2024). Specifically, we extend recent research on the potential use of LLMs in student assessments (González-Calatayud et al. 2021; Wetzler et al. 2024). Unlike some prior evidence, our findings suggest promising results for implementing GPT models in higher education assessment tasks. However, we emphasize that consistency is not perfect, as deviations between human and GPT scores persist in this type of subjective evaluation. Therefore, we do not support the idea of completely replacing human graders but instead propose using GPT models as a complementary tool for teaching assessments.

Scholarly literature has extensively discussed strategies to reduce bias and enhance fairness in grading (e.g., Kates et al. 2023; Malouff, Emmerton, and Schutte 2013; Peter, Karst, and Bonefeld 2024). Our findings provide pathways and replicable tools for implementing GPT models as a second graders in subjective assessments, thereby contributing to the reduction of grading biases. Our correlation analysis and Bland-Altman plots are particularly useful for identifying cases that may require re-assessment. Rather than a drawback, the inconsistencies between human and GPT scoring can be leveraged to flag exams that may require further review, enhancing the fairness in assessments. Thus, the promise of GPT lies not in replacing human graders but in advancing the equity of evaluations.

We also compared different GPT models and settings, a critical exercise given the varying costs of implementation. GPT-4o-mini is significantly more cost-effective than GPT-4o, approximately 33 times cheaper for processing input and 25 times cheaper for generating responses (Context 2025). While GPT-4o exhibits slightly higher consistency with human graders, GPT-4o-mini remains effective for detecting deviations and offers a cost-efficient alternative, particularly at lower temperature settings. This cost-effectiveness also makes GPT-4o-mini a good option for image-to-text transcription. It is also possible that instructors may require direct submission of digital answers to avoid the first step of the process.

This study opens several avenues for future research. First, we demonstrated how GPT can be used for transcribing and scoring handwritten essay-type questions as a second grader. Future studies could extend this approach to other forms of assessment, such as project-based assignments or longer essay responses, since students in this study were constrained to only half-page answers. Second, we did not explore variations in consistency based on students' background characteristics. Predicting the differences between human and GPT scores by considering factors such as ethnicity, gender, or discipline could reveal specific sources of bias or skewness, enabling educators to address them and promote fairness. Finally, our data came from an introductory sociology course, and it would be valuable to examine GPT performance in other disciplines with varying levels of subjectivity.

**Data Availability statement**

Configuration files and Python scripts can be found at https://github.com/KamToAzr/image-transcription-and-grading.git.

**References**

Barros, Amon, Ajnesh Prasad, and Martyna Śliwa. 2023. "Generative Artificial Intelligence and Academia: Implication for Research, Teaching and Service." *Management Learning* 54(5):597–604.

Birkelund, Johan. 2014. "The Lunch Effect. Can It Result in Biased Grading at Universities?" UiT Norges arktiske universitet.

Bland, J., and D. Altman. 1999. "Measuring Agreement in Method Comparison Studies." *Statistical Methods in Medical Research* 8(2):135–60.

Chin, Mark J., David M. Quinn, Tasminda K. Dhaliwal, and Virginia S. Lovison. 2020. "Bias in the Air: A Nationwide Exploration of Teachers' Implicit Racial Attitudes, Aggregate Bias, and Student Outcomes." *Educational Researcher* 49(8):566–78.

Context. 2025. "Understand and Compare GPT-4o Mini vs. GPT-4o." Retrieved (https://context.ai/compare/gpt-4o-mini/gpt-4o).

Danziger, Shai, Jonathan Levav, and Liora Avnaim-Pesso. 2011. "Extraneous Factors in Judicial Decisions." *Proceedings of the National Academy of Sciences of the United States of America* 108(17):6889–92.

Digital Education Council. 2024. *Digital Education Council Global AI Student Survey 2024: AI or Not AI: What Students Want.*

González-Calatayud, Víctor, Paz Prendes-Espinosa, and Rosabel Roig-Vila. 2021. "Artificial Intelligence for Student Assessment: A Systematic Review." *Applied Sciences 2021, Vol. 11,*

*Page 5467* 11(12):5467.

Grivokostopoulou, Foteini, Isidoros Perikos, and Ioannis Hatzilygeroudis. 2017. "An Educational System for Learning Search Algorithms and Automatically Assessing Student Performance." *International Journal of Artificial Intelligence in Education* 27(1):207–40.

Jukiewicz, Marcin. 2024. "The Future of Grading Programming Assignments in Education: The Role of ChatGPT in Automating the Assessment and Feedback Process." *Thinking Skills and Creativity* 52:101522.

Kates, Sean, Tine Paulsen, Sidak Yntiso, and Joshua A. Tucker. 2023. "Bridging the Grade Gap: Reducing Assessment Bias in a Multi-Grader Class." *Political Analysis* 31(4):642–50.

Klein, Joseph, and Liat Pat El. 2003. "Impairment of Teacher Efficiency during Extended Sessions of Test Correction." *European Journal of Teacher Education* 26(3):379–92.

Lee, Dabae, and Sheunghyun Yeo. 2022. "Developing an AI-Based Chatbot for Practicing Responsive Teaching in Mathematics." *Computers & Education* 191:104646.

Levy, Dan, and Angela Pérez Albertos. 2024. *Teaching Effectively with ChatGPT.*

van der Loo, Mark P. J. 2014. "The Stringdist Package for Approximate String Matching." *R Journal* 6(1):111–22.

Malouff, John M., Ashley J. Emmerton, and Nicola S. Schutte. 2013. "The Risk of a Halo Bias as a Reason to Keep Students Anonymous During Grading." *Teaching of Psychology* 40(3):233–37.

Malouff, John M., Sarah J. Stein, Lodewicka N. Bothma, Kimberley Coulter, and Ashley J. Emmerton. 2014. "Preventing Halo Bias in Grading the Work of University Students." *Cogent Psychology* 1(1):988937.

Mandernach, B. Jean, and Rick Holbeck. 2016. "Teaching Online: Where Do Faculty Spend Their Time?" *Online Journal of Distance Learning Administration* 19(4).

McCarthy, J., M. L. Minsky, N. Rochester, and C. E. Shannon. 1955. "A Proposal for the Darmouth Summer Research Project on Artificial Intelligence." *AI Magazine* 27(4):12–14.

Olivos, Francisco, and Minhui Liu. 2024. "ChatGPTest: Opportunities and Cautionary Tales of Utilizing AI for Questionnaire Pretesting." *Field Methods*.

OpenAI. 2024a. "ChatGPT-4o [Dec 2025 Version] [Large Language Model]."

OpenAI. 2024b. "GPT-4o Mini: Advancing Cost-Efficient Intelligence." Retrieved (https://openai.com/index/gpt-4o-mini-advancing-cost-efficient-intelligence/).

OpenAI. 2025. "Create Chat Completion." Retrieved January 10, 2025 (https://platform.openai.com/docs/api-reference/chat/create).

Peter, Sophia, Karina Karst, and Meike Bonefeld. 2024. "Objective Assessment Criteria Reduce the Influence of Judgmental Bias on Grading." *Frontiers in Education* 9:1386016.

Popenici, Stefan A. D., and Sharon Kerr. 2017. "Exploring the Impact of Artificial Intelligence on Teaching and Learning in Higher Education." *Research and Practice in Technology Enhanced Learning* 12(1):1–13.

Protivínský, Tomáš, and Daniel Münich. 2018. "Gender Bias in Teachers' Grading: What Is in the Grade." *Studies in Educational Evaluation* 59:141–49.

Ratten, Vanessa, and Paul Jones. 2023. "Generative Artificial Intelligence (ChatGPT): Implications for Management Educators." *The International Journal of Management Education* 21(3):100857.

Srihari, Sargur, Jim Collins, Rohini Srihari, Harish Srinivasan, Shravya Shetty, and Janina Brutt-Griffler. 2008. "Automatic Scoring of Short Handwritten Essays in Reading Comprehension Tests." *Artificial Intelligence* 172(2–3):300–324.

Stoltz, Dustin S., and Marshall A. Taylor. 2024. *Mapping Texts : Computational Text Analysis for the Social Sciences*. New York: Oxford University Press.

Wang, Shan, Fang Wang, Zhen Zhu, Jingxuan Wang, Tam Tran, and Zhao Du. 2024. "Artificial Intelligence in Education: A Systematic Literature Review." *Expert Systems with Applications* 252:124167.

Wetzler, Elizabeth L., Kenneth S. Cassidy, Margaret J. Jones, Chelsea R. Frazier, Nickalous A. Korbut, Chelsea M. Sims, Shari S. Bowen, and Michael Wood. 2024. "Grading the Graders: Comparing Generative AI and Human Assessment in Essay Evaluation." *Teaching of Psychology*.

**Prompt 1a**

*This is a photo of a student's answer to a question on a mid-term exam for an Introduction to Sociology course at a university in [ANONYMIZED]. Please transcribe the student's answer exactly as written. If any word is unclear or unreadable, skip it and continue with the transcription. If the student did not answer the question, return only [BLANK]. Ensure the output is plain text containing solely the transcription without any additional text, explanations, formatting, or tags. Do not include headers, footers, introductory phrases, markdown, HTML tags, or any other annotations. For example, do not add elements like Transcription:, Here is the transcription of the student's answer:, or similar phrases.*

[base64 coded of the student answer]

Figure 1. Prompt for literal transcription.

**Prompt 2a**

*This is a photo of a student's answer to a question on a mid-term exam for an Introduction to Sociology course at a university in [ANONYMIZED]. Please transcribe the student's answer to the best of your ability. If any word is unclear or unreadable, please transcribe it with your best guess. If you make a guess, do not enclose the guessed words in parentheses, brackets, etc. If the student did not answer the question, return only [BLANK]. Ensure the output is plain text containing solely the transcription without any additional text, explanations, formatting, or tags. Do not include headers, footers, introductory phrases, markdown, HTML tags, or any other annotations. For example, do not add elements like Transcription:, Here is the transcription of the student's answer:, or similar phrases.*

[base64 coded of the student answer]

Figure 2. Prompt for literal transcriptions utilizing the GPT models' best guess.

<div style="border: 1px solid black; padding: 10px;">

**Prompt 1b**

***Context:*** *You are an instructor in [ANONYMIZED], and your students have completed an exam comprising six questions. This exam aims to explore various sociological theories and applications. Your task is to evaluate the responses using the scoring guidelines provided.*

***Objective:*** *Assess each response using your own knowledge and calculate the total score out of a maximum of 10 points.*

***Instructions for Grading:***

  • *Evaluate each response individually.*
  • *Score from 0 to 1 point each for questions 1 and 2, and from 0 to 2 points for questions 3 through 6.*
  • *Your scores can cover the complete decimal range from .1 to .9*
  • *Provide a score for each question and calculate the total score.*
  • *Do not add comments or feedback; focus solely on scoring.*

***Task:***

  • *Score each question based on your own knowledge and the provided student responses.*
  • *Please return your scores in one string, each score separated by an underscore. Example: score1_score2_score3_score4_score5_score6.*
      • *Just return the scores in this format and nothing else.*

</div>

Figure 3. Scoring prompt utilizing GPT models' sociological knowledge.

<div style="border: 1px solid black; padding: 10px;">

**Prompt 2b**

***Context:*** *You are an instructor in [ANONYMIZED], and your students have completed an exam comprising six questions. This exam aims to explore various sociological theories and applications. Your task is to evaluate the responses using the scoring guidelines provided.*

***Objective:*** *Assess each response using the correct answers below and calculate the total score out of a maximum of 10 points.*

***Exam Questions and correct answers:*** *(See supplementary material).*

***Instructions for Grading:***

  • *Evaluate each response individually.*
  • *Score from 0 to 1 point each for questions 1 and 2, and from 0 to 2 points for questions 3 through 6.*
  • *Your scores can cover the complete decimal range from .1 to .9*
  • *Provide a score for each question and calculate the total score.*
  • *Do not add comments or feedback; focus solely on scoring.*

***Task:***

  • *Score each question based on your own knowledge and the provided student responses.*
  • *Please return your scores in one string, each score separated by an underscore. Example: score1_score2_score3_score4_score5_score6.*
  • *Just return the scores in this format and nothing else.*

</div>

Figure 4. Scoring prompt providing template answers.

Table 1. Similarity Index of human and GPT transcriptions.

| Question | Prompt | GPT Model | Mean Cosine | SD Cosine |
|---|---|---|---|---|
| 1 | 1a | GPT-4o-mini | 0.9811 | 0.0255 |
| | | GPT-4o | 0.9748 | 0.0310 |
| | 2a | GPT-4o-mini | 0.9827 | 0.0232 |
| | | GPT-4o | 0.9755 | 0.0308 |
| 2 | 1a | GPT-4o-mini | 0.9765 | 0.0257 |
| | | GPT-4o | 0.9722 | 0.0372 |
| | 2a | GPT-4o-mini | 0.9787 | 0.0264 |
| | | GPT-4o | 0.9728 | 0.0363 |
| 3 | 1a | GPT-4o-mini | 0.9836 | 0.0183 |
| | | GPT-4o | 0.9768 | 0.0227 |
| | 2a | GPT-4o-mini | 0.9840 | 0.0192 |
| | | GPT-4o | 0.9807 | 0.0198 |
| 4 | 1a | GPT-4o-mini | 0.9780 | 0.0243 |
| | | GPT-4o | 0.9764 | 0.0259 |
| | 2a | GPT-4o-mini | 0.9828 | 0.0193 |
| | | GPT-4o | 0.9782 | 0.0240 |
| 5 | 1a | GPT-4o-mini | 0.9793 | 0.0292 |
| | | GPT-4o | 0.9803 | 0.0189 |
| | 2a | GPT-4o-mini | 0.9822 | 0.0266 |
| | | GPT-4o | 0.9799 | 0.0265 |
| 6 | 1a | GPT-4o-mini | 0.9807 | 0.0203 |
| | | GPT-4o | 0.9791 | 0.0175 |
| | 2a | GPT-4o-mini | 0.9849 | 0.0132 |
| | | GPT-4o | 0.9810 | 0.0159 |
| Mean | 1a | GPT-4o-mini | 0.9798 | 0.0241 |
| | | GPT-4o | 0.9765 | 0.0265 |
| | 2a | GPT-4o-mini | 0.9825 | 0.0218 |
| | | GPT-4o | 0.9700 | 0.0264 |

*Note:* Prompt 1a instructs literal transcription; prompt 2a instructs GPT models' best guess.

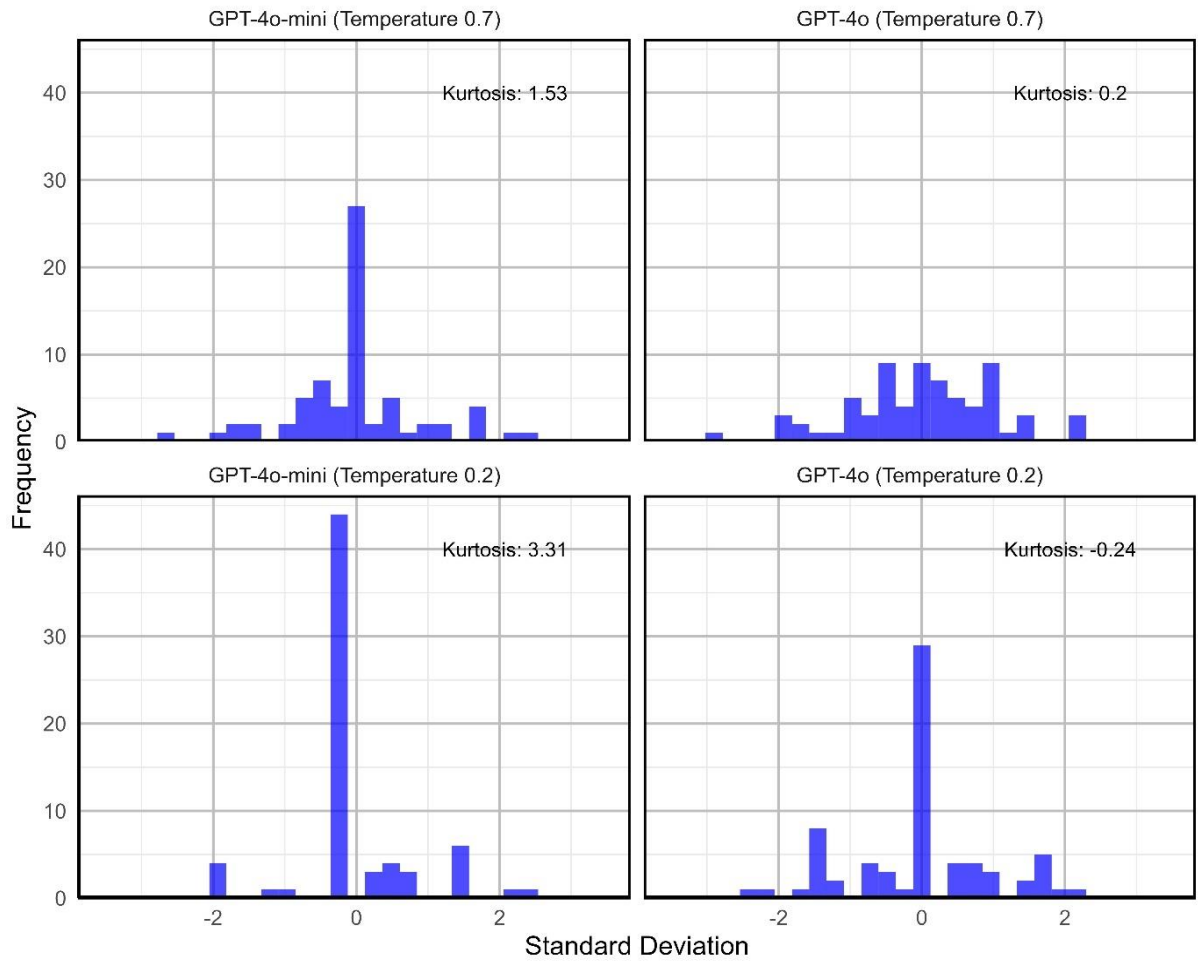

Figure 5. Standard deviation of 100 scoring tasks without prompting template for each setting.

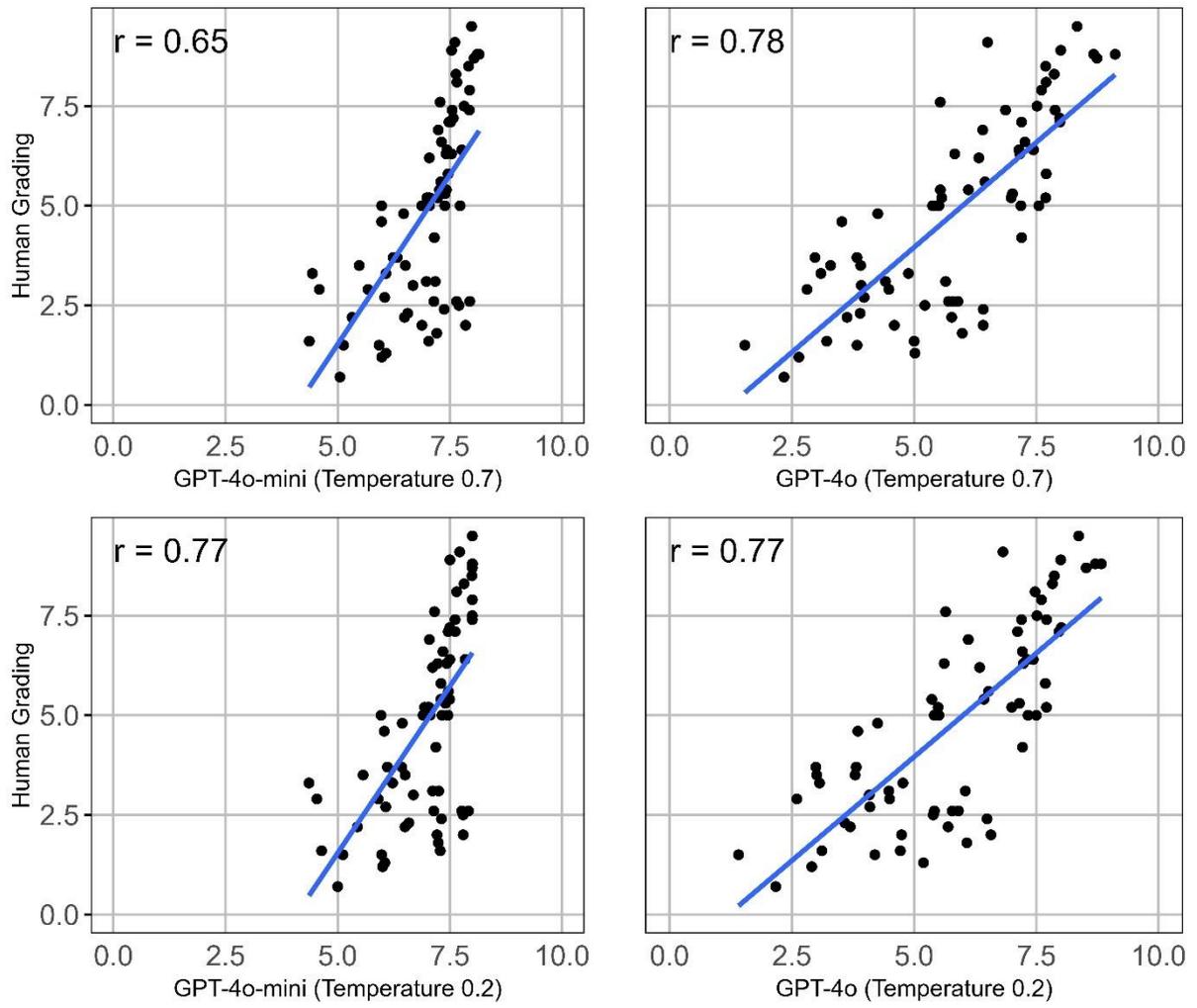

Figure 6. Correlation between total human and GPT scoring without template answers for each setting.

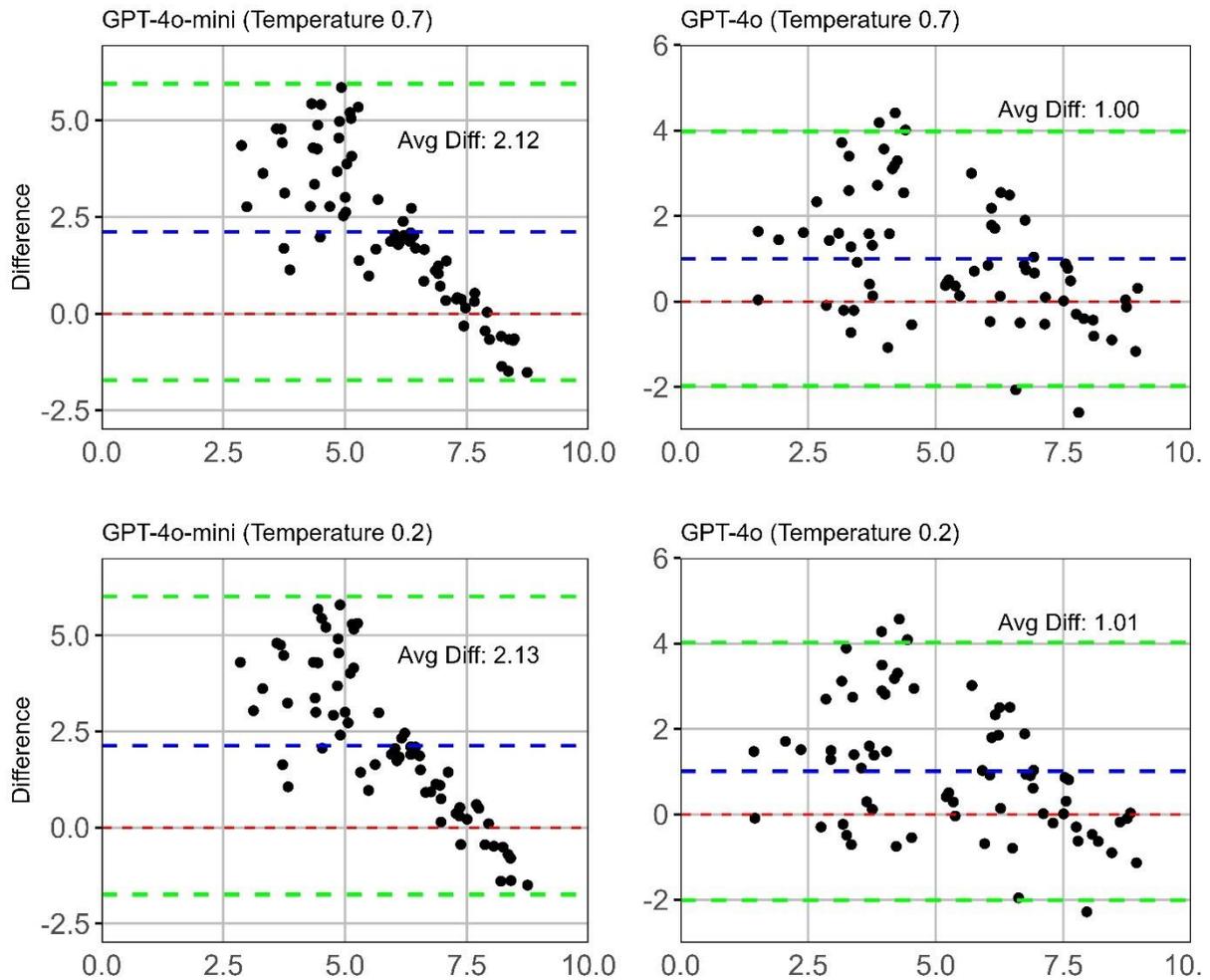

Figure 7. Bland-Altman graph for average scores and differences between human scores and GPT without prompting template answers.

*Note:* The average scores represent the mean of the human and GPT scores for each student. The difference is calculated by subtracting human scores from GPT scores for each student. The blue dashed line represents the mean difference (average bias) between the two sets of scores. The green dashed lines show the upper and lower limits of agreement, calculated as the mean difference plus and minus 1.96 times the standard deviation of the differences, respectively.

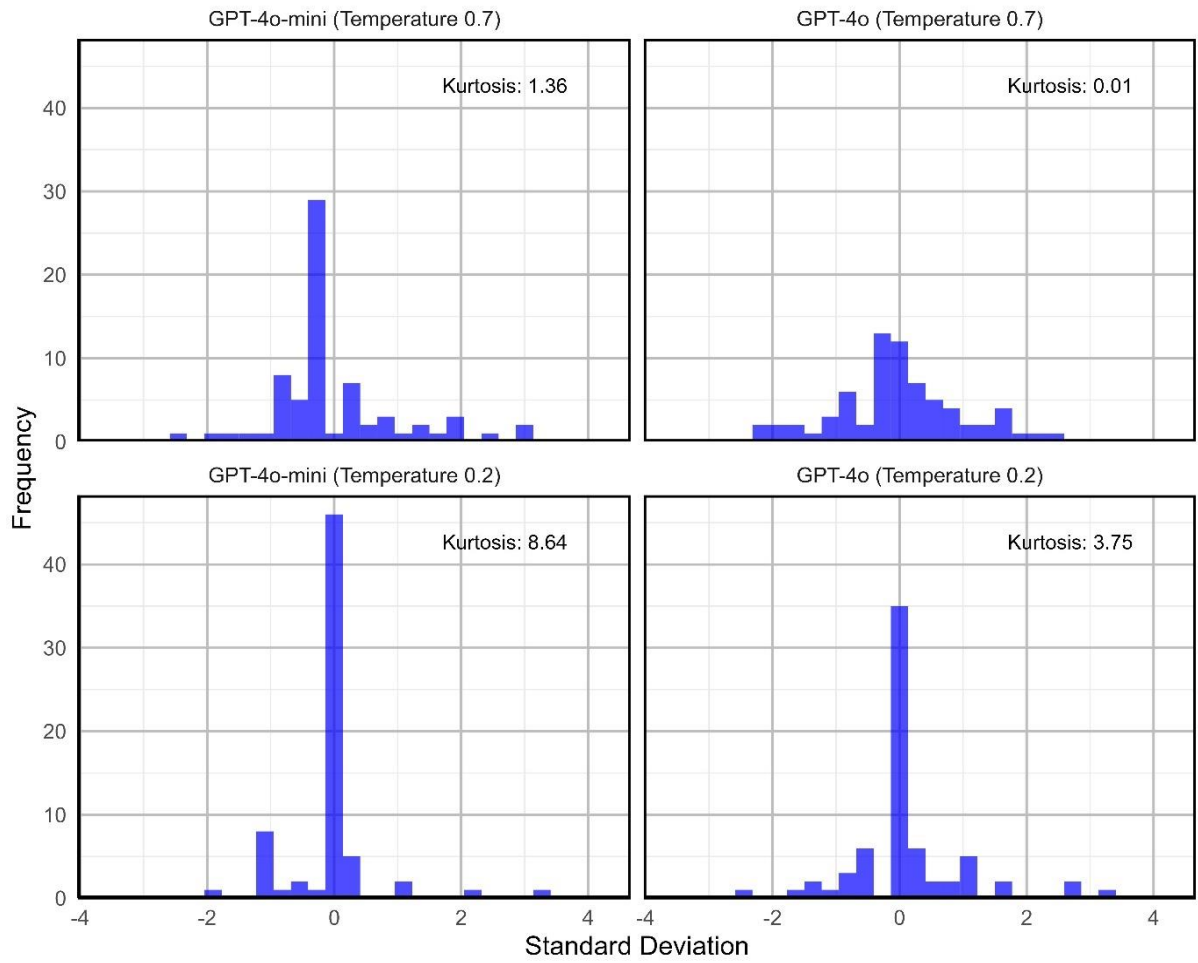

Figure 8. Standard deviation of 100 scoring tasks prompting template answers for each setting.

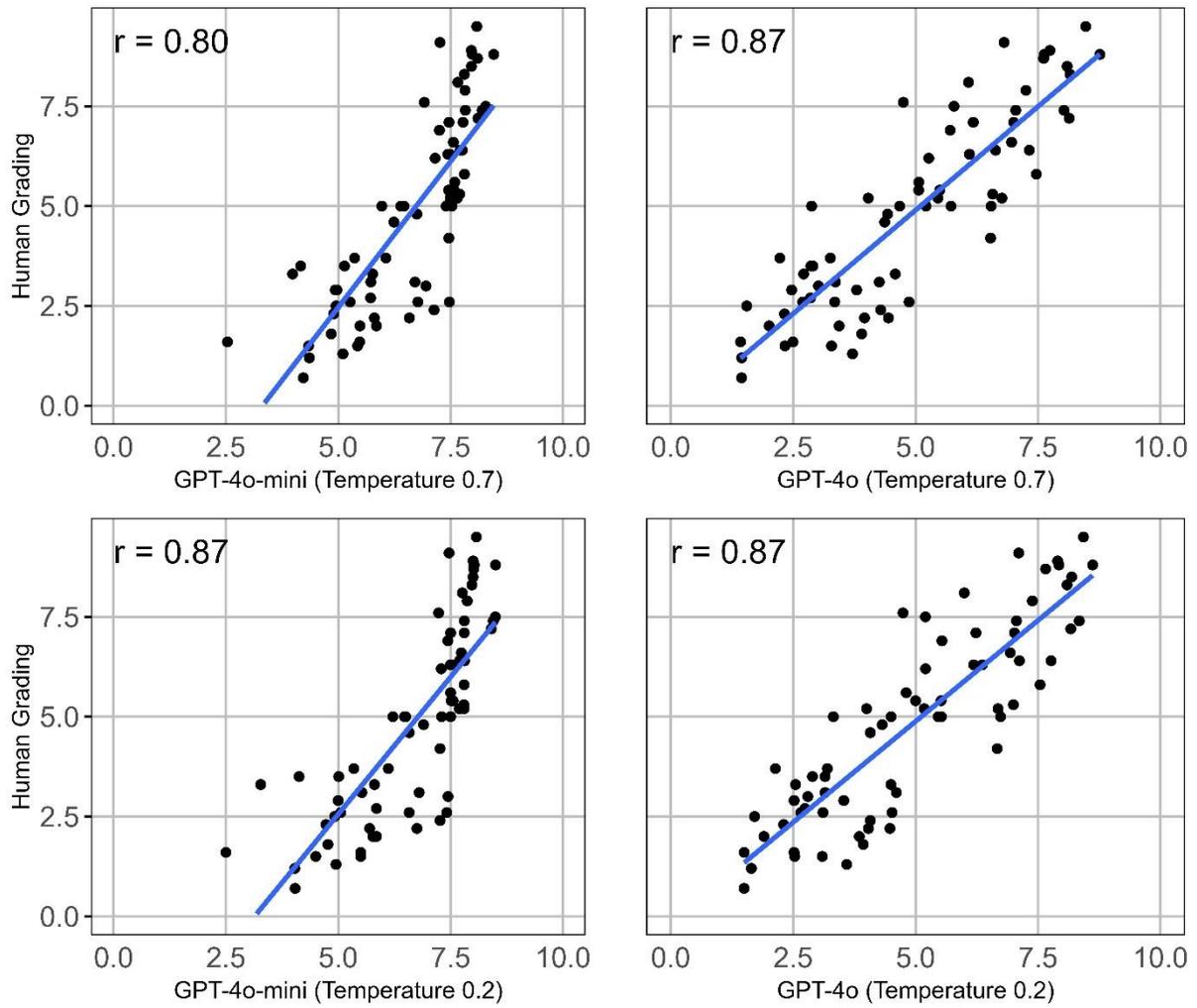

Figure 9. Correlation between total human and GPT scoring with template answers for each setting.

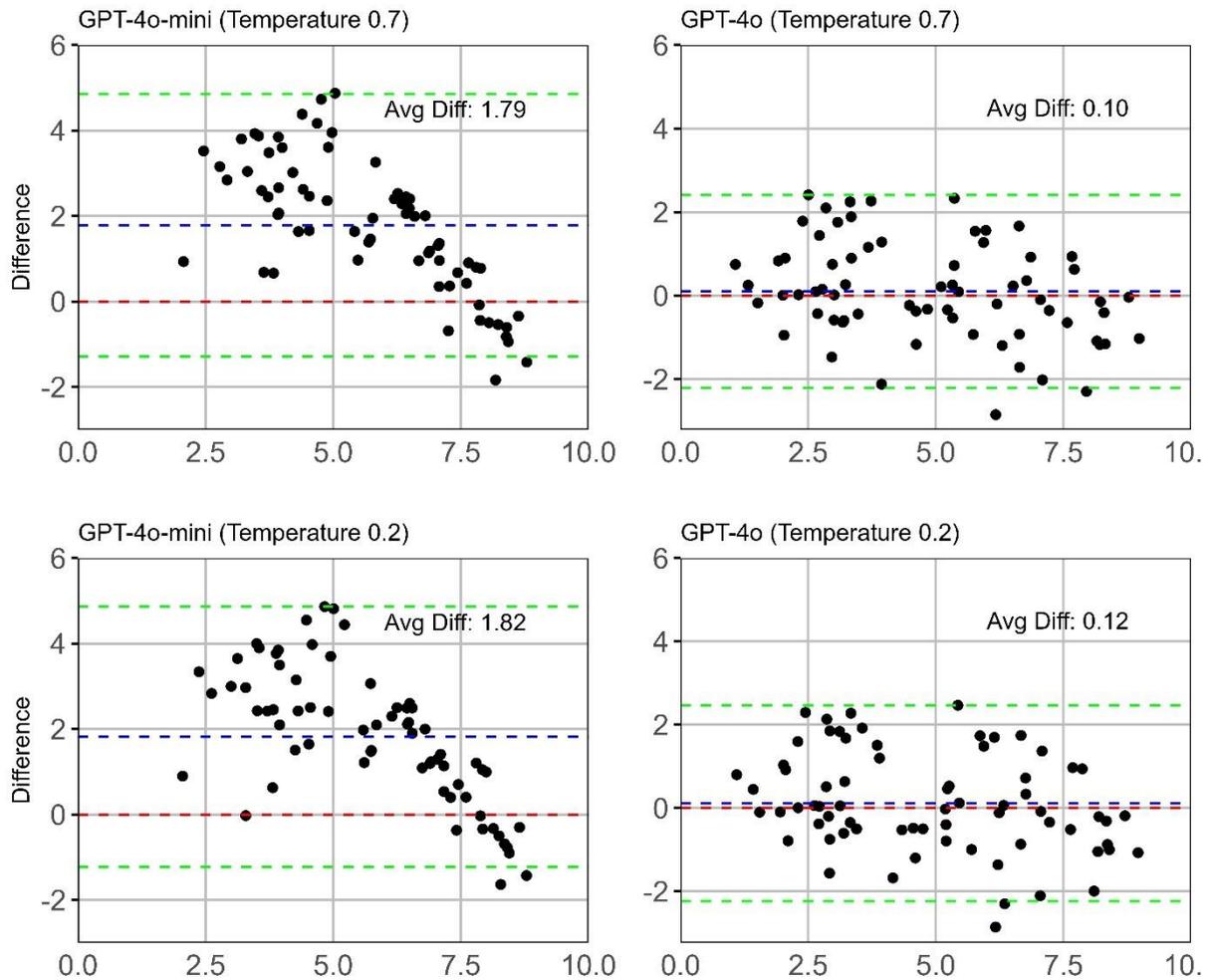

Figure 10. Bland-Altman graph for average scores and differences between human scores and GPT prompting template answers.

*Note*: The average scores represent the mean of the human and GPT scores for each student. The difference is calculated by subtracting human scores from GPT scores for each student. The blue dashed line represents the mean difference (average bias) between the two sets of scores. The green dashed lines show the upper and lower limits of agreement, calculated as the mean difference plus and minus 1.96 times the standard deviation of the differences, respectively.

Table 2. Pearson's correlation and average difference between human and GPT scores for each question utilizing prompt 2.

| Question | Grader | 4o-mini Temperature 0.7 | | 4o Temperature 0.7 | | 4o-mini Temperature 0.2 | | 4o Temperature 0.2 | |
|---|---|---|---|---|---|---|---|---|---|
| | | r | Diff. | r | Diff. | r | Diff. | r | Diff. |
| 1 | A | 0.68 | 0.22 | 0.76 | 0.18 | 0.67 | 0.23 | 0.78 | 0.18 |
| 2 | A | 0.87 | 0.26 | 0.83 | 0.15 | 0.84 | 0.26 | 0.81 | 0.16 |
| 3 | A | 0.57 | 0.57 | 0.64 | 0.32 | 0.55 | 0.58 | 0.60 | 0.34 |
| 4 | A | 0.79 | 0.30 | 0.85 | 0.35 | 0.77 | 0.29 | 0.82 | 0.35 |
| 5 | B | 0.59 | 0.69 | 0.70 | 0.53 | 0.60 | 0.67 | 0.70 | 0.53 |
| 6 | C | 0.45 | 0.63 | 0.76 | 0.35 | 0.45 | 0.62 | 0.75 | 0.36 |

**Assessing instructor-AI cooperation for grading essay-type questions in an introductory**

**sociology course**

### Template answers

*1. Can Asians be considered a social group? Justify your answer.*

No. A social group consists of two or more people who identify with and interact with one another. Under that definition, Asians are a category because they share a common status. However, although they know others share the same status, most are strangers to one another and do not necessarily interact.

*2. There are three major theoretical perspectives in sociology: structural functionalism, social conflict, and symbolic interactionism. Explain the difference between them in terms of their level of analysis and what each perspective focuses on.*

Structural functionalism and social conflict perspectives adopt a macro-level analysis and focus on social structures that shape society as a whole. In contrast, symbolic interactionism adopts a micro-level approach that focuses on social interactions in specific situations.

*3. Although people often view crime as the result of free choice or personal failings, why does a sociological approach argue that crime is shaped by society?*

Crime is a specific category of deviance, where a society's formally enacted criminal law has been violated. As a type of deviance, there are three social foundations that suggests that society shapes crime. First, crime varies according to cultural norms. No thought or action is inherently criminal; it becomes crime only in relation to specific norms. Second, people become criminal as others defined them that way. Whether a behaviour defines us as criminal depends on how other perceive, define, and respond to it (e.g., jury). Third, how societies set norms and how they define rule breaking both involve social power. Norms and their application reflect social inequality. The powerless is more likely to be consider as a criminal than the powerful.

*4. Charles Darwin's groundbreaking 1859 study of evolution led people to believe that human behavior was instinctive, simply a product of our "nature." However, social scientists in the 20th century debunked this idea. Explain how the concept of socialization contradicts the evolutionist argument that behavior is simply our nature.*

In explaining the behavioral differences between individuals from different societies, scientists originally believed they were explained by biological differences—namely, nature. However, social scientists have demonstrated that behavior is not instinctive but learned, a process that sociologists refer to as socialization. Socialization is the lifelong social experience through which people develop their human potential and learn culture. The cultural norms and values, which constrain our behavior, are acquired during the process of socialization. Thus, the concept of socialization contradicts the evolutionist argument by suggesting that what we consider nurture might actually be the true nature of humans.

*5. According to Peter Berger, how does the sociological perspective differ from other ways of understanding human behavior, such as those offered by psychology, political science, or legal frameworks?*

Berger argues that the sociological perspective differs from other disciplines like psychology or political science because it seeks to go beyond individual-level explanations (such as psychological motives) or official interpretations (such as those provided by political institutions). This approach involves looking "beyond the immediately given and publicly approved interpretations", which are not reflected in the everyday understanding of a given phenomenon.

6. *Marvin Harris's cultural ecology framework challenges the assumption that religious practices are purely symbolic or superstitious. Explain why cow worship in India makes sense from an economic perspective.*

The main assumption of Harris' framework is that culture is an adaptation that arises from environmental and ecological constraints. Thus, cow worship makes sense in India because cows supply long-lasting and sustainable resources necessary Indian rural households in an ecological environment that is prone to heavy droughts and monsoon seasons.